\documentclass[10pt,twocolumn,letterpaper]{article}

\usepackage{cvpr}              % To produce the CAMERA-READY version
\usepackage{graphicx}
\usepackage{amsmath}
\usepackage{amssymb}
\usepackage{booktabs}

\usepackage{amsmath,amsfonts,bm}

\def\eqref#1{equation~\ref{#1}}
\def\1{\bm{1}}

\def\mF{{\bm{F}}}

\def\mM{{\bm{M}}}

\def\mP{{\bm{P}}}

\def\mW{{\bm{W}}}

\DeclareMathAlphabet{\mathsfit}{\encodingdefault}{\sfdefault}{m}{sl}
\SetMathAlphabet{\mathsfit}{bold}{\encodingdefault}{\sfdefault}{bx}{n}

\usepackage{multirow}

\usepackage[pagebackref,breaklinks,colorlinks]{hyperref}

\usepackage[capitalize]{cleveref}
\crefname{section}{Sec.}{Secs.}
\Crefname{section}{Section}{Sections}
\Crefname{table}{Table}{Tables}
\crefname{table}{Tab.}{Tabs.}

\def\cvprPaperID{*****} % *** Enter the CVPR Paper ID here
\def\confName{CVPR}
\def\confYear{2022}

\begin{document}

\title{Team VI-I2R Technical Report on EPIC-KITCHENS-100 Unsupervised Domain Adaptation Challenge for Action Recognition 2022}

\author{Yi Cheng$^1$, Dongyun Lin$^1$, Fen Fang$^1$, Hao Xuan Woon$^{1,2}$, Qianli Xu$^1$, Ying Sun$^1$ \\
$^1$Institute for Infocomm Research, Agency for Science, Technology and Research, Singapore \\
$^2$National University of Singapore\\
{\tt\small \{cheng\_yi,lin\_dongyun,fang\_fen,qxu,suny\}@i2r.a-star.edu.sg, haoxuan.woon@u.nus.edu}
}

\maketitle

%%%%%%%%% ABSTRACT
\begin{abstract}
  In this report, we present the technical details of our submission to the EPIC-KITCHENS-100 Unsupervised Domain Adaptation (UDA) Challenge for Action Recognition 2022. This task aims to adapt an action recognition model trained on a labeled source domain to an unlabeled target domain. To achieve this goal, we propose an action-aware domain adaptation framework that leverages the prior knowledge induced from the action recognition task during the adaptation. Specifically, we disentangle the source features into action-relevant features and action-irrelevant features using the learned action classifier and then align the target features with the action-relevant features. To further improve the action prediction performance, we exploit the verb-noun co-occurrence matrix to constrain and refine the action predictions. Our final submission achieved the first place in terms of top-1 action recognition accuracy. 
  
\end{abstract}

%%%%%%%%% BODY TEXT
\section{Introduction}
\label{sec:intro}

The EPIC-KITCHENS-100 dataset is a large-scale video dataset, capturing daily cooking activities in different kitchens using head-mounted cameras~\cite{damen2022ijcv}. It mainly contains fine-grained actions involving extensive hand object interactions, and each action in the dataset is defined by the combination of a verb and a noun. The Unsupervised Domain Adaptation (UDA) for Action Recognition Challenge aims to learn an action recognition model on a labeled source domain and generalize it to an unlabeled target domain. It has attracted increasing attention from the community as it can significantly alleviate the annotation burden when applying a trained model to other unannotated datasets. 

Compared with UDA for image-based tasks, such as image classification and object detection, UDA for video-based tasks is more challenging as both spatial features and temporal dynamics should be aligned during the adaptation. In the task of UDA for Action Recognition, adversarial learning is the dominant approach that aims to learn domain-invariant features for action recognition~\cite{chen2019temporal}. 
Although rapid progress has been made, these methods have one intrinsic limitation, \ie, they directly align source and target features which may degrade the performance of action recognition. It is known that the essence of action recognition is to learn discriminative action-relevant features. Similarly, for UDA for action recognition, it is desirable to ensure that the target features are discriminative enough for correct prediction. However, as the source video features contain both action-relevant and action-irrelevant features, directly aligning source and target features would introduce extra noise and reduce the discriminability of learned features. Therefore, it is important to align the target features with only action-relevant source features. 

To address this limitation, we propose to leverage the prior knowledge generated from the action recognition task for video domain adaptation. Specifically, the source feature is first disentangled into action-relevant and action-irrelevant source features using the action classifier learned on the source data, and then the target feature is aligned with the action-relevant source feature. In this manner, the model can learn discriminative domain-invariant features for action recognition. 
Besides, as each action class is defined as the combination of a verb and a noun, some combinations may be invalid (\eg, rinse \& table). We exploit the verb-noun co-occurrence matrix generated from the source domain to constrain and refine the action predictions.

%-------------------------------------------------------------------------
\section{Our Approach}
\begin{figure*}[t]
  \centering
  \includegraphics[width=1.0\linewidth]{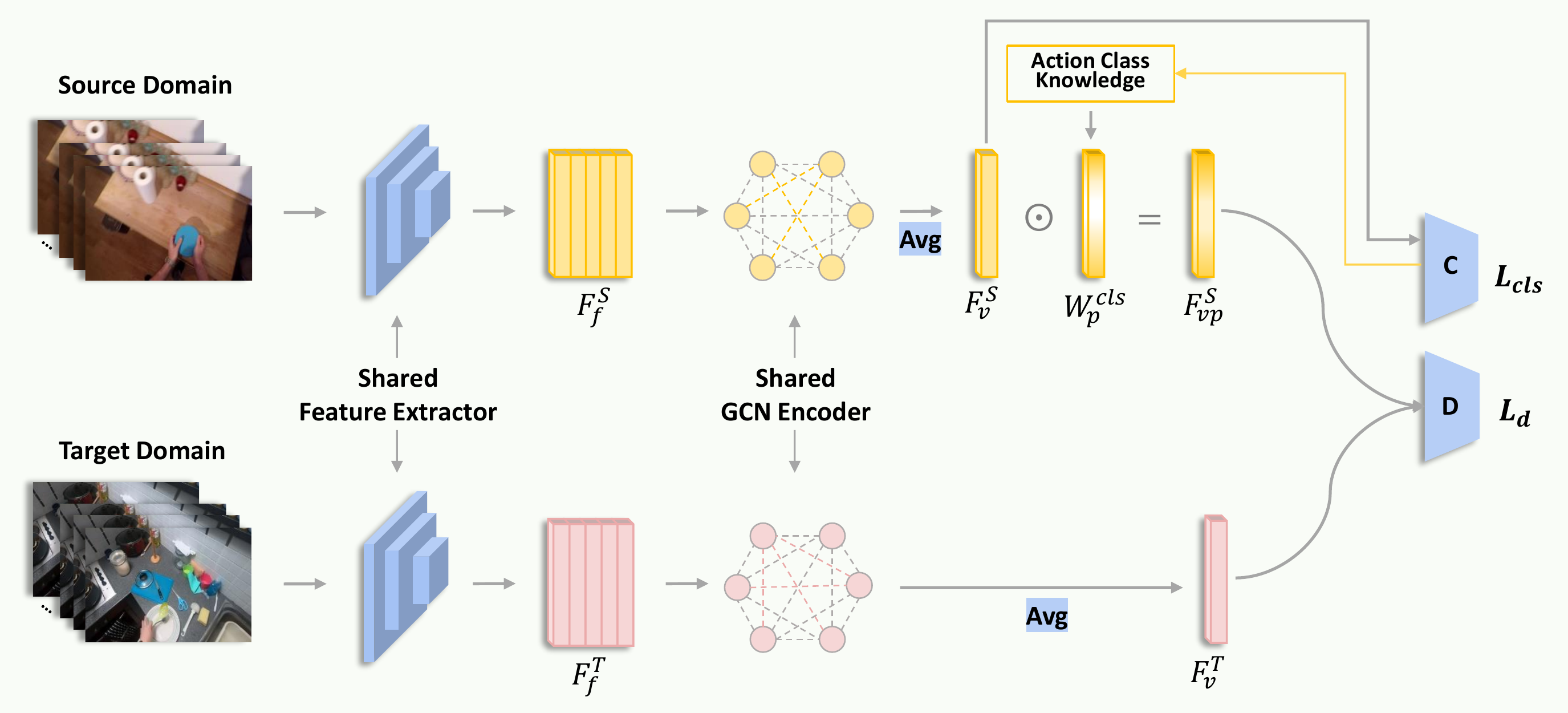}
  \caption{Overall architecture of the proposed framework. First, the frame-level source/target features ($\mF_f^S$/$\mF_f^T$) are extracted from video frames using a pre-trained feature extractor. Then, the extracted features are passed to a Graph Convolutional Networks (GCNs), followed by an average operation, to generate the video-level source/target feature ($\mF_v^S$/ $\mF_v^T$). Next, $\mF_v^S$ is passed to the action classifier, and then the action-discriminative features $\mF_{vp}^S$ are generated using the action class knowledge learned by the action classifier. Lastly, we align $\mF_v^T$ with $\mF_{vp}^S$ for video domain adaptation. $L_{cls}$ and $L_d$ denote action classification loss and domain classification loss, respectively. This figure is best viewed in color.} 
\label{fig:architect}
\end{figure*}

In this section, we describe the technical details of our proposed approach. As illustrated in Fig.~\ref{fig:architect}, the overall framework mainly contains two stages: video representation learning and action-aware domain adaptation. We will describe each stage in the following subsections. 

%------------------------
% \vspace{3mm}
%------------------------

\subsection{Video Representation Learning}
To learn a robust video feature representation that can generalize across domains for action recognition, it is essential to mine the intrinsic temporal relations within videos. Therefore, we design a video representation learning module that consists of a pre-trained feature extractor for frame feature encoding and a GCN encoder for temporal relation modeling. 

\noindent\textbf{Feature extractors.} 
To generate powerful feature representations from the input videos, we explore SlowFast~\cite{slowfast2019iccv}, a model based on the 3D Convolutional Neural Network, to extract features from the input video frames. The extracted features are used to generate the video-level features for video domain adaptation.  

\noindent\textbf{GCN encoder.} 
As the feature extractor maps individual video frames into the corresponding frame-level features, it does not fully explore the intrinsic temporal structure in videos. Therefore, we apply a fully-connected GCN encoder to model the temporal relations between different video frames. Concretely, we first embed the extracted features from both the source and target domains into the graph space using an FC layer, where the dimension of output features is $D$. Then, the GCN encoder takes the embedded features as input and outputs a sequence of frame-level features containing rich temporal relation information. Then, we perform average pooling on the output features to generate the video-level feature representations $\mF_v^S$ and $\mF_v^T$.

\subsection{Action-aware Domain Adaptation}
In the task of UDA for action recognition, it is essential to ensure that the shared feature embeddings across domains are discriminative enough for action classification. Therefore, we propose disentangling the action-relevant features from the holistic source features to enable the action-aware alignment with target features. 

Grad-CAM~\cite{gradcam2018wacv} is a popular technique to identify the discriminative features for CNN-based classification models~\cite{cheng2022entropy, lin2022multi}. 
It has been explored in~\cite{lin2021KBScam,lin2021cam,lin2020cam} that weights of the learned classifier with respect to the ground-truth class can help to identify the critical features for correct class prediction. Motivated by this observation, we propose to use weights of the learned action classifier for the ground-truth action class to generate the action-relevant features that are discriminative for action classification. Concretely, with the video-level source feature $\mF_v^S$ and the weights of learned action classifier $\mW_p^{cls}$ for ground-truth class $p$, the action-relevant feature is computed as:
 \begin{equation}
     \mF_{vp}^S = \mW_p^{cls} \odot \mF_v^S, 
 \end{equation}
where $\odot$ is the Hadamard product, $\mF_{vp}^S$ is the action-relevant feature containing critical information for action classification. 

After obtaining the action-relevant features $\mF_{vp}^S$ from the source domain, features from the target domain are aligned with $\mF_{vp}^S$ using a domain classifier to discriminate whether the sample is from the source or target domain. Following~\cite{chen2019temporal}, we insert a gradient layer between the domain classifier and the main model for gradient back-propagation. As shown in Fig.~\ref{fig:architect}, the overall framework is optimized using two loss functions: the action classification loss $L_{cls}$ using source action labels and the domain classification loss $L_d$. 

\subsection{Verb-noun Co-occurrence Prior}
During inference, the video-level target feature $\mF_{v}^T$ is passed to the learned action classifier for action prediction. Since each action class is defined as a combination of a verb and a noun, we design two classification branches: one for predicting the verb probabilities $\mP_{V}$ and the other for predicting the noun probabilities $\mP_{N}$. Therefore, the action probabilities are computed as:
 \begin{equation}
     \mP_{A} = \mP_{V}\mP_{N}^T.
 \end{equation}
As mentioned in Section~\ref{sec:intro}, some action classes are invalid because certain verb classes and noun classes are incompatible, such as rinse and table. Therefore, we propose to utilize the co-occurrence of verb and noun as prior knowledge to refine the final predictions on target samples. Concretely, we compute $\mM$, the co-occurrence matrix of verb and noun, from the statistics of the source domain, where $\mM_{i,j}$ denotes the number of co-occurrence times of the $i$-th verb class and the $j$-th noun class. With the assumption that action classes never appearing in the source domain are highly likely to be invalid, we refine the action probabilities on target samples by reducing the probabilities of invalid action classes:
 \begin{equation}
     \mP_{A}^{'} = \mP_{A} \odot \mM^{'}, \mM^{'}=
     \begin{cases}
        1, & \text{if } \mM_{i,j}>0,\\
        0.01, & \text{otherwise.}
    \end{cases}
 \end{equation}

%------------------------------------------------------------------------
\section{Experiments}
\label{sec:exp}

\subsection{Implementation Details}
\noindent\textbf{Feature extractors.}
We train three variants of the SlowFast~\cite{slowfast2019iccv}, including SlowFast with ResNet50, SlowFast with ResNet101, and SlowOnly (using only the slow path in SlowFast) with ResNet50. For each of the three variants, the model is trained for 60 epochs using synchronized SGD training as in~\cite{slowfast2019iccv}. The input number of frames are set as $32$ and $8$ for the fast and slow paths, respectively. The batch size is set as $64$. During feature extraction, we extract the features from the last convolutional layer and apply average pooling to generate the frame-level feature representations. The feature dimension for SlowOnly-ResNet50 is $2048$, while the feature dimensions for SlowFast-ResNet50 and SlowFast-ResNet101 are $2304$.

\noindent\textbf{Action-aware domain adaptation.} 
We follow the guidelines posted by the challenges to train the action-aware domain adaptation model. The model is first trained on the validation set for algorithm validation and hyper-parameters tuning. Then, the model is retrained on the training set using the selected hyper-parameters. Finally, the model is applied to predict the action labels of target samples in the testing set, followed by a refinement on the action predictions, to generate the final results. During training, the parameters of the feature extractors are fixed, while the other parameters are learned using an initial learning rate at $3 \times 10^{-3}$. The model is trained for $60$ epochs, and the learning rate is multiplied by $0.1$ after $30$ and $45$ epochs. We empirically set the dimension of embedded feature vectors  as $D=512$.

\subsection{Results}
Table~\ref{tab:ablation} demonstrates the recognition performance on the target validation set using the RGB and Flow features extracted from the pre-trained SlowOnly-ResNet50 model. It is observed that by leveraging the action-relevant information from the learned action classifier, the performance of model can be improved by 1.72\% in terms of top-1 action accuracy on the validation set. Moreover, the refinement using verb-noun co-occurrence prior information can further improve the top-1 action accuracy by 0.59\%.
As the EPIC-KITCHENS-100 dataset is highly imbalanced with many tail classes containing very few training samples, the learned action classifier may not be informative enough for the tail classes. Therefore, we expect a higher performance gain on a balanced dataset with enough training samples.

%----------------------------
\begin{table}[ht]
\caption{The comparison of model performance on the EPIC-KITCHENS-100 validation set. ``Baseline" denotes the general domain adaptation without using the information from the learned action classifier. ``Baseline+ADA" denotes the Action-aware Domain Adaptation (ADA) proposed in the report. ``Baseline+ADA+AF" denotes our proposed method including the Action Refinement (AF). The best performance is marked in bold.}
\centering
\scalebox{0.76}{
\begin{tabular}{c|c|c|c|c|c|c}
\hline
\multirow{2}{*}{Method} & \multicolumn{3}{c|}{Top-1 Accuracy (\%)} & \multicolumn{3}{c}{Top-5 Accuracy (\%)} \\ \cline{2-7} 
 & Verb & Noun & Action & Verb & Noun & Action \\ \hline
Baseline & 50.33  &  34.30 & 22.63 &79.75  & 56.15  & 48.41 \\ \hline
Baseline+ADA & 52.75 & 34.76 & 24.35 & 81.33 & 58.08 & 50.57 \\ \hline
Baseline+ADA+AF & 52.75 & 34.76 & \textbf{24.94} & 81.33 & 58.08 & \textbf{51.62} \\ \hline

\end{tabular}
}
\label{tab:ablation}
\end{table}
%----------------------------

%----------------------------
\begin{table}[ht]
\caption{The performance of different models on the EPIC-KITCHENS-100 validation set.}
\centering
\scalebox{0.65}{
\begin{tabular}{c|c|c|c|c|c|c|c}
\hline
\multirow{2}{*}{Feature extractor} &\multirow{2}{*}{Input} & \multicolumn{3}{c|}{Top-1 Accuracy (\%)} & \multicolumn{3}{c}{Top-5 Accuracy (\%)} \\ \cline{3-8} 
& & Verb & Noun & Action & Verb & Noun & Action \\ \hline
SlowOnly (R50) & RGB+Flow & 52.75 & 34.76 & 24.94 & 81.33 & 58.08 & 51.62 \\ \hline
SlowFast (R50) & RGB & 49.55 & 33.35 & 23.01 & 80.57 & 56.03 & 49.82 \\ \hline
SlowFast (R101) & RGB & 46.79 & 34.81 & 23.24 & 78.24 & 56.02 & 49.70 \\ \hline
\end{tabular}
}
\label{tab:ensemble}
\end{table}
%----------------------------

%----------------------------
\begin{table*}[t]
\caption{The final results of UDA for domain adaptation on the EPIC-KITCHENS-100 test set.}
\centering
\scalebox{0.85}{
\begin{tabular}{c|c|c|c|c|c|c}
\hline
\multirow{2}{*}{Method} & \multicolumn{3}{c|}{Top-1 Accuracy (\%)} & \multicolumn{3}{c}{Top-5 Accuracy (\%)} \\ \cline{2-7} 
 & Verb & Noun & Action & Verb & Noun & Action \\ \hline
Ensemble & 57.89  &  40.07 & \textbf{30.12} &83.48  & 64.19  & 48.10 \\ \hline
\end{tabular}
}
\label{tab:test}
\end{table*}
%----------------------------

\subsection{Model Ensemble}
As model ensemble helps exploit the complementary nature of predictions from different models~\cite{Sun2020TeamVT}, we ensemble the results from models shown in Table~\ref{tab:ensemble}. These models are trained on features extracted using the three variants of the SlowFast~\cite{slowfast2019iccv} action recognition model. To further improve the performance, we also ensemble the results from HC-VDA~\cite{Cheng2021TeamVT} which leverages the hand bounding boxes to generate hand-centric features for video domain adaptation. Following~\cite{huang2021towards}, we first calculate the action predictions for each model and then aggregate the results in terms of action probabilities. The final results on the test set are shown in Table~\ref{tab:test}, and it ranks first in terms of the top-1 action accuracy in the EPIC-KITCHENS-100 UDA Challenge for Action Recognition 2022.

%------------------------------------------------------------------------
\section{Conclusion}
In this report, we describe the technical details of our approach to the EPIC-KITCHENS-100 UDA Challenge for Action Recognition 2022. To leverage the action-relevant information that are invariant across domains, we propose an action-aware domain adaptation framework for action recognition. To the best of our knowledge, this is the first work to exploit the prior knowledge induced from the learned action classifier in the task of UDA for action recognition. Moreover, we utilize the verb-noun co-occurrence matrix computed from the source domain data to refine the action predictions. With further performance increase from the model ensemble, our final submission ranks first on the leaderboard in terms of top-1 action recognition accuracy.

%------------------------------------------------------------------------
\section*{Acknowledgments}
We would like to thank Dr. Joo Hwee Lim for his continuous support and useful guidance. This research is supported by the Agency for Science, Technology and Research (A*STAR) under its AME Programmatic Funding Scheme (Project \#A18A2b0046). 

%%%%%%%%% REFERENCES
{\small
\bibliographystyle{ieee_fullname}
\bibliography{egbib}
}

\end{document}